\documentclass[]{llncs}
\usepackage[T1]{fontenc}
\usepackage{caption}
\usepackage{amsmath,amssymb,amsfonts}
\usepackage{algorithmic}
\usepackage{graphicx}
\usepackage{textcomp}
\usepackage{multirow}
\usepackage{xcolor}
\usepackage{booktabs}
\usepackage{caption}
\usepackage{amsmath} 
\usepackage{orcidlink}
\usepackage{authblk}

\usepackage{subcaption}
\usepackage{graphicx}
\begin{document}
\title{	Dynamic Uncertainty Learning with Noisy Correspondence for Text-Based Person Search}

%
%
\author{Zequn Xie  \inst{1}\orcidID{0009-0007-5021-7565} \and
Haoming Ji   \inst{2}\orcidID{0009-0008-9771-2413}\and
Chengxuan Li   \inst{3} \and
Lingwei Meng   \inst{4} }

\authorrunning{}
%
\institute{Zhejiang University \and
Beijing University of Posts Telecommunications \and 
Beijing Forestry University \and
Northwest Normal University 
\\
\email{zqxie@zju.edu.cn,jihaoming@bupt.edu.cn,lcx0904@bjfu.edu.cn,1251610681@qq.com}  
}

\maketitle   
\begin{abstract}

Text-to-image person search aims to identify an individual based on a text description. To reduce data collection costs, large-scale text-image datasets are created from co-occurrence pairs found online. However, this can introduce noise, particularly mismatched pairs, which degrade retrieval performance. Existing methods often focus on negative samples, which amplify this noise. To address these issues, we propose the Dynamic Uncertainty and Relational Alignment (DURA) framework, which includes the Key Feature Selector (KFS) and a new loss function, Dynamic Softmax Hinge Loss (DSH-Loss). KFS captures and models noise uncertainty, improving retrieval reliability. The bidirectional evidence from cross-modal similarity is modeled as a Dirichlet distribution, enhancing adaptability to noisy data. DSH adjusts the difficulty of negative samples to improve robustness in noisy environments. Our experiments on three datasets show that the method offers strong noise resistance and improves retrieval performance in both low- and high-noise scenarios.

\end{abstract}
\section{Introduction}

Person Re-identification (Re-ID) is a key task in computer vision, important for applications like intelligent video surveillance, urban security, and smart retail \cite{10657514}. Text-to-image person search aims to identify a person based on textual descriptions. However, current Re-ID methods rely on specific images of individuals, which may not be available in real-world emergency situations. In these cases, eyewitness descriptions can be the only information. To address this, text-based person search is used to identify individuals from image collections based on text queries.

Text-to-image person search is a sub-domain of both image-text retrieval \cite{2,3,4} and image-based person Re-ID \cite{5,6,7}. Text descriptions offer an intuitive way to express a person’s features, often easier to access than images. This has led to growing interest in the field, with applications in personal photo searches and public safety. The main challenge is accurately measuring the similarity between text descriptions and images to retrieve the correct individual from the image database based on the provided text query.

To address these issues, current methods \cite{Bai_2023} use techniques to increase the similarity of positive text-image pairs while reducing that of negative pairs. Approaches like common representation learning \cite{8} and similarity learning \cite{9} have been applied. While these methods show promise, many depend on large, well-annotated datasets. To reduce data collection costs, co-occurring text-image pairs are often gathered from online sources, creating a large, cost-effective cross-modal dataset. However, such data often contains noise, known as noisy correspondence, which weakens the reliability of cross-modal relationships and affects retrieval performance. This problem is especially severe with hinge-based triplet ranking loss involving hard negatives \cite{8,9}, where the hard learning approach becomes more vulnerable to noise, lowering retrieval accuracy, as shown in our experiments (e.g., Table \ref{tab1}).

Noisy correspondence is closely related to learning with noisy labels \cite{10,11}, a concept studied in classification tasks. Various methods, like co-teaching \cite{11} and robust loss functions \cite{10}, have been developed to address noisy labels. However, noisy correspondence involves misalignments between cross-modal pairs \cite{9577762,10204951,Li2023DCELDC,10024790}, which differs from noisy labels in classification. Traditional robust learning techniques are inadequate for noisy correspondence, as it involves uncertainty at the instance level rather than the category level. As a result, noisy correspondence is much more complex than noisy labels, as the number of instances usually exceeds the number of categories by far.

To address these challenges, we propose the Dynamic Uncertainty and Relational Alignment (DURA) framework, designed for text-to-image person search in noisy environments. DURA introduces a new Cross-modal Evidential Learning (CEL) method to identify and handle uncertainty caused by noise, helping to isolate unreliable pairs and improve data reliability. To deal with unreliable hard negatives, the Dynamic Softmax Hinge (DSH) mechanism increases the difficulty of negative samples during training, reducing the impact of noisy correspondence. By combining CEL and DSH, DURA effectively distinguishes and uses both clean and noisy data. CEL models bidirectional evidence as a Dirichlet distribution based on cross-modal similarity, addressing uncertainty from noisy correspondence. This evidence helps classify the training data into clean and noisy subsets. Finally, DURA uses CEL and DSH loss functions to train the model, applying positive learning to the clean data and negative learning to the noisy data. In conclusion, the main contributions of this work are as follows:

\begin{itemize}
    \item We propose Dynamic Uncertainty and Relational Alignment (DURA) framework to provide trusted retrieval in an effective and efficient way. Our DURA could be directly applied to robustly learn with noisy correspondence for Text-to-Image person search.
    \item A Dynamic Softmax Hinge Loss (DSH-Loss) is proposed to mitigate the adverse effects of unreliability caused by noisy correspondence. Specifically, DSH smoothly increases the difficulty of negative samples during training to improve the robustness against noisy correspondence.
    \item Extensive experiments verify that the proposed method improves the robustness against noisy correspondence, especially the high noise rate. Moreover, we provide insightful analysis that the learned uncertainty could reduce the negative impacts of noisy correspondences, improving the robustness.
\end{itemize}

\section{Related Work}

\subsection{Text-to-Image Person Search}

The concept of \textbf{Text-to-Image Person Search} was introduced by Li \textit{et al.} \cite{1} with the \textbf{CUHK-PEDES} dataset. Early methods used VGG and LSTM \cite{1,13} for feature extraction, aligning image-text features via matching loss functions. Later works \cite{15,16,17} employed ResNet and BERT to enhance representations and design advanced matching losses. Current approaches are categorized into \textbf{global-matching} \cite{18,19,20,21}, which align features in a unified space, and \textbf{local-matching} \cite{22,23,24,25}, which focuses on fine-grained alignments like body parts and text elements for better retrieval.{A recent work  \cite{xie2025chatdriventextgenerationinteraction} explored a novel approach involving chat-driven text generation and interaction for person retrieval, opening new avenues for incorporating interactive and generative methods into the task.
Despite progress, most methods assume perfect training pairs, which is unrealistic due to noise. To address this, Yang \textit{et al.} \cite{26} proposed a robust learning method leveraging isovariant similarity consistency for noisy data. However, it depends on high-quality anchor points, and poor selection can lead to misclassification. }

\subsection{Uncertainty-based Learning}

Many deep models\cite{27} have been used with great success in applications, but they use deterministic predictions and lack the ability to assess their output uncertainty. To overcome this challenge, many researches\cite{franchi2023makebnnsimplestrategy,jaskari2022uncertaintyawaredeeplearningmethods,ahmed2024scalableefficientmethodsuncertainty} proposed methods to estimate the output uncertainty of the models. Kendall \textit{et al.} \cite{28} proposed the Bayesian deep learning framework combining input-related intrinsic uncertainty with knowledge-based uncertainty, focusing on the application of the model in per-pixel semantic segmentation and deep regression tasks. Chen \textit{et al.} \cite{29} simultaneously modelled the output uncertainty of the models by considering multi-granularity uncertainty for both coarse-grained and fine-grained image retrieval, integrating uncertainty modelling and uncertainty regularisation to improve recall and enhance the retrieval process. In contrast, our approach focuses more on modelling uncertainty across schema correspondences and aims to achieve robustness and reliability in text-to-image person search.

\section{Methodology}

\begin{figure*}[htbp]

\centering
\includegraphics[scale=0.55]{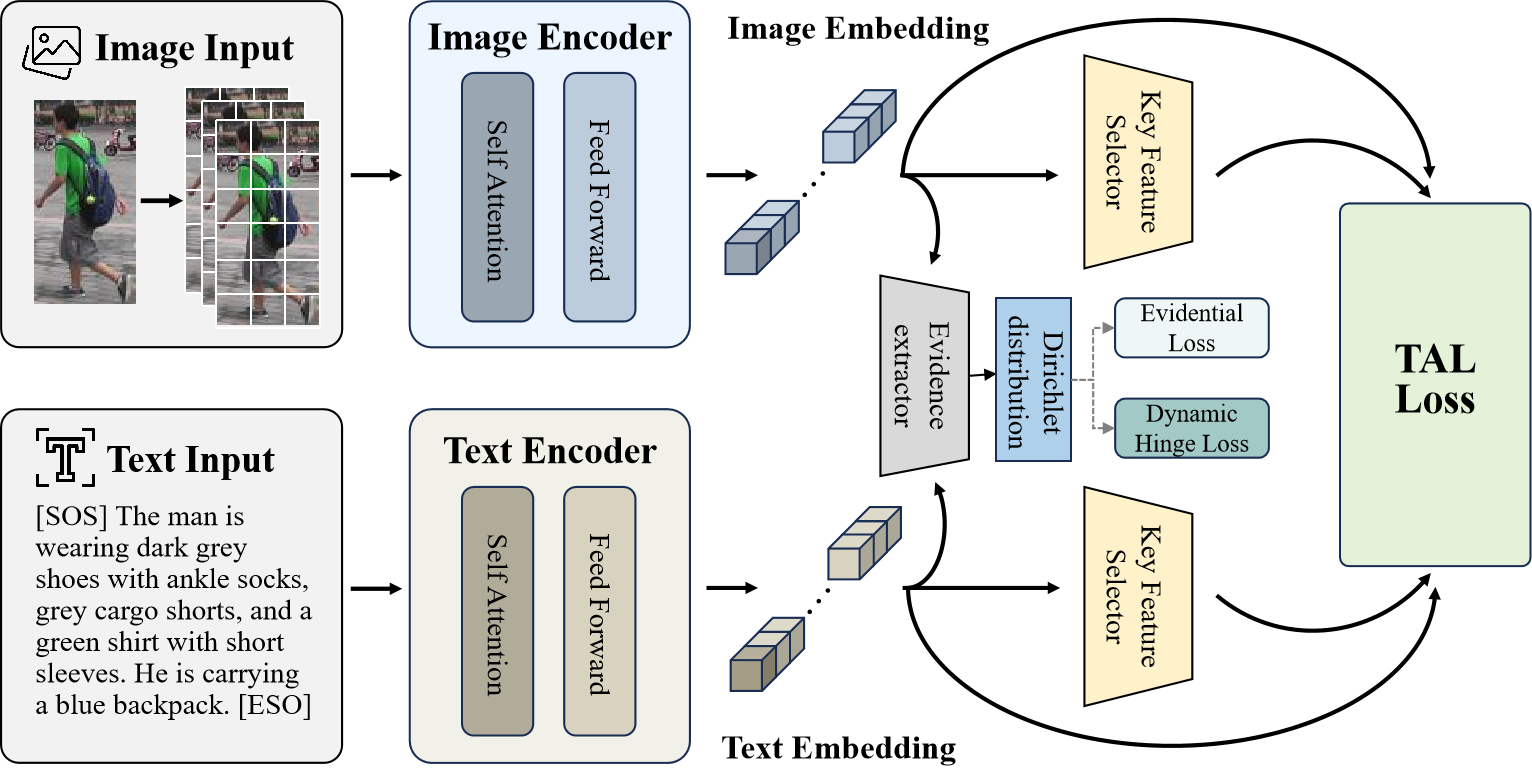}  
\captionsetup{font=small,justification=justified,singlelinecheck=false}
\caption{\textbf{The overview of our Dynamic Uncertainty and Relational Alignment framework (DURA).illustrating feature extraction, evidence-based uncertainty modeling, and evidence-guided training with TAL, CEL, DSH, and KFS.}}
\label{fig1}
 
\end{figure*}

In this section, we presents the proposed Dynamic Uncertainty and Relational Alignment (DURA) framework, which incorporates the Key Feature Selector (KFS) module and a novel loss function, the Dynamic Softmax Hinge Loss (DSH-Loss), to achieve robust cross-modal retrieval.The overview of DURA is illustrated in Fig. \ref{fig1} and the details are discussed in the following subsections.

\subsection{Feature Extraction with Dual-Encodert}

Our framework uses a dual-encoder architecture for feature extraction, utilizing pre-trained encoders for both image and text to ensure strong cross-modal alignment. Following the success of CLIP in aligning visual and textual embeddings, we initialize both the image and text encoders with the full CLIP architecture, improving the model's ability to extract semantically aligned features from both modalities.

Formally, the gallery set is represented as $V = { (I_i, y^p_i, y^v_i) }{i=1}^{N_v}$, and the corresponding textual description set is $T = { (T_i, y^v_i) }{i=1}^{N_t}$, where $N_v$ and $N_t$ denote the number of images and texts, respectively. The label $y^p_i \in Y^p = { 1, \dots, C }$ is the class label (person identity), with $C$ being the total number of identities, and $y^v_i \in Y^v = { 1, \dots, N_v }$ is the image label. The paired image-text dataset for TIPeID is defined as $P = { (I_i, T_i, y^v_i, y^p_i) }_{i=1}^N$, where each image-text pair shares the same image label $y^v_i$ and class label $y^p_i$.

\textbf{Image Encoder:} Given an input image \( I \in \mathbb{R}^{H \times W \times C} \), We use a CLIP pre-trained Vision Transformer (ViT) to extract image embeddings. An input image \( I \in \mathbb{R}^{H \times W \times C} \) is divided into \( N = \frac{H \times W}{P^2} \) non-overlapping patches, where \( P \) is the patch size. Each patch is flattened and transformed into a token representation \(\{f_v^i\}_{i=1}^N\), with positional embeddings added and a special \([CLS]\) token prepended. This tokenized sequence is processed through \( L \)-layer transformer blocks with self-attention, and the \([CLS]\) token output \( f_v^{\text{cls}} \) captures the global image representation. A linear projection maps \( f_v^{\text{cls}} \) to a shared image-text embedding space for alignment with text embeddings.

\textbf{Text Encoder:} Given an input text \( T_i \), We use a CLIP pre-trained transformer-based text encoder. Input text \( T_i \) is tokenized using byte pair encoding (BPE) into a sequence of subword tokens. Special tokens \([\text{SOS}]\) and \([\text{EOS}]\) are added to mark the boundaries, resulting in a sequence \(\{f_t^{\text{sos}}, f_t^1, \dots, f_t^N, f_t^{\text{eos}}\}\). This sequence is processed through \( L \)-layer transformer blocks with self-attention, and the \([\text{EOS}]\) token output \( f_t^{\text{eos}} \) captures the global text representation. A linear projection maps \( f_t^{\text{eos}} \) to the shared image-text embedding space, enabling alignment with image embeddings.

To measure the similarity between an image-text pair \((I_i, T_j)\), we utilize the global features represented by the \text{[CLS]} and \text{[EOS]} tokens. The global embedding similarity \( S_{ij} \) is computed using cosine similarity, defined as:

\begin{equation}
S_{ij}=\frac{f_v^{\mathrm{cls}^\top}\cdot f_t^{\mathrm{Eos}}}{\|f_v^{\mathrm{cls}}\|\|f_t^{\mathrm{Eos}}\|}
\end{equation}

Here, \( f_t^{\mathrm{eos}} \) represents the global text embedding derived from the \text{[EOS]} token, while \( f_v^{\mathrm{cls}} \) denotes the global image embedding obtained from the \text{[CLS]} token. These global embeddings are used to compute the similarity \( S_{ij} \) between the image and text in the shared multimodal embedding space. Additionally, \( \{ f_t^j \}_{j=1}^{N} \) and \( \{ f_v^j \}_{j=1}^{N} \) correspond to the local features extracted from the \( N \) word tokens in the text \( T_i \) and the \( N \) patches in the image \( I_i \), respectively, providing fine-grained contextual representations.

\subsection{ Key Feature Selector}

While global embeddings obtained from \([CLS]\) and \([EOS]\) tokens capture high-level similarities between image-text pairs, they often overlook subtle, fine-grained details essential for accurate text-to-image person search. To address this issue, we introduce a \textbf{Key Feature Selector (KFS)} module that enhances the discriminative power of the learned representations by incorporating informative local features.

We begin by applying L2 normalization to both visual and textual features:
$
\hat{f_v^{\text{i}}} = \text{L2Norm}(f_v^{\text{i}}) \quad \text{and} \quad \hat{f_t^{\text{i}}} = \text{L2Norm}(f_t^{\text{i}})
$
This normalization step ensures consistent magnitudes across features, improving their stability and reliability, especially under noisy conditions.

Next, we refine features using Max-K pooling, which selects the top \( k \) values and averages them. This process emphasizes the most discriminative components, allowing the model to focus on the critical cues that distinguish one identity from another.

Before pooling, we further enhance feature representations using a combination of MLP, FC, and a Squeeze-and-Excitation (SE) layer at the MLP’s output. The MLP and FC layers transform the input features into a richer, more expressive space, while the SE layer adaptively recalibrates channel-wise feature responses, highlighting informative dimensions and suppressing less relevant ones.

Formally, the visual and textual features \(F_v^{i}\) and \(F_t^{i}\) are computed as:
\begin{align}
F_v^{i} &= \text{MaxPool}\bigl(\text{SE}(\text{MLP}(V_i^{s})) + \text{FC}(f_v^{i})\bigr), \\
F_t^{i} &= \text{MaxPool}\bigl(\text{SE}(\text{MLP}(T_i^{s})) + \text{FC}(f_t^{i})\bigr)
\end{align}
By integrating global and local perspectives within DURA, the KFS module ensures a balanced representation that captures both high-level alignment and fine-grained distinctions. This leads to improved robustness and accuracy in text-to-image person search, particularly in complex and noisy retrieval scenarios.

\subsection{Uncertainty Modeling}

In this section, we model cross-modal uncertainty using the Dempster-Shafer Theory of Evidence, guided by the principles of Subjective Logic. Consider a mini-batch of \( N \) image-text pairs. For each pair \((I_i, T_j)\), we first compute a similarity score \( S_{ij} \). To translate this score into evidence, we apply an evidence extractor \( f(\cdot) \):
\begin{equation}
e_{ij}=f(S_{ij})=\exp\left(\tanh\left(S_{ij}/\tau\right)\right),
\label{eq:evidence_function}
\end{equation}

where \( 0 < \tau < 1 \) is a scaling parameter. Thus, for a given visual query \( I_i \), the evidence vector \( \mathbf{e}_{i \to T_i} \) can be extracted from the cross-modal similarities using Equation~\eqref{eq:evidence_function}, i.e.,
$
    \mathbf{e}_{i \to T_i} = [e_{i1}, e_{i2}, \dots, e_{iN}].
$
Similarly, the evidence vector \( \mathbf{e}_{T_i \to i} \) for a given textual query \( T_i \) can be obtained as:
$
    \mathbf{e}_{T_i \to i} = [e_{1i}, e_{2i}, \dots, e_{Ni}].
$

Subjective Logic assigns a belief mass to each query and an overall uncertainty mass based on the collected cross-modal evidence (e.g., \( \mathbf{e}_{i \to T_i} \) and \( \mathbf{e}_{T_i \to i} \)), which can be defined as:
\begin{equation}
    b_{ij} = \frac{e_{ij}}{L_i} \quad and \quad u_i = \frac{N}{L_i},
\end{equation}

where:
$
    L_i = \sum_{j=1}^N (e_{ij} + 1)\quad and \quad u_i + \sum_{j=1}^N b_{ij} = 1.
$The term \( L_i \) represents the strength of the Dirichlet distribution, while the belief mass assignment \( \mathbf{b}_i = [b_{i1}, b_{i2}, \dots, b_{iN}] \) corresponds to subjective opinions derived from the Dirichlet distribution with parameters \( \boldsymbol{\alpha}_i = [\alpha_{i1}, \alpha_{i2}, \dots, \alpha_{iN}] \), where:
$
    \alpha_{ij} = e_{ij} + 1.
$

Intuitively, cross-modal retrieval is analogous to classifying instances (i.e., pairs), where the query similarity aligns with the probability assignment. The Dirichlet distribution, parametrized by the evidence, represents the density of these probability assignments. Thus, \( \boldsymbol{\alpha}_i \) models second-order probabilities and uncertainties. The density function is defined as:
\begin{equation}
    D(\mathbf{p}_i | \boldsymbol{\alpha}_i) =
    \begin{cases}
        \frac{1}{B(\boldsymbol{\alpha}_i)} \prod_{j=1}^N p_{ij}^{\alpha_{ij} - 1}, & \text{for } \, \mathbf{p}_i \in \mathcal{S}_N, \\
        0, & \text{otherwise},
    \end{cases}
\end{equation}

where \( \mathbf{p}_i \in \mathbb{R}^N \) are the query probabilities, \( B(\boldsymbol{\alpha}_i) \) is the \( N \)-dimensional multinomial beta function, and \( \mathcal{S}_N \) denotes the \( N \)-dimensional unit simplex .

In this manner, the evidence extracted from cross-modal similarities informs a second-order distribution (the Dirichlet), representing both the likelihood of each potential match and the associated uncertainty. This approach ensures a more robust handling of noisy and uncertain retrieval scenarios.

\subsection{Cross-modal Evidential Learning}
Following the approach of \cite{qin2022deep}, we treat cross-modal retrieval as a \(K\)-way classification task, where each query is expected to match its corresponding target. Let \(\mathbf{y}_i\) be a \(K\)-dimensional one-hot vector indicating the correct match for query \(i\). The model produces evidence \(\boldsymbol{\alpha}_i = [\alpha_{i1}, \dots, \alpha_{iK}]\) that parameterizes a Dirichlet distribution, modeling both the probability assignments and their associated uncertainty.

To align the model’s probability estimates \(\mathbb{E}[p_{ij}] = \frac{\alpha_{ij}}{L_i}\) (with \( L_i = \sum_{k=1}^K \alpha_{ik} \)) to the ground truth \(\mathbf{y}_i\), we employ the following mean-squared loss over the Dirichlet distribution:
\[
\mathcal{L}_m(\boldsymbol{\alpha}_i, \mathbf{y}_i) 
= \sum_{j=1}^K \left[ \left( y_{ij} - \frac{\alpha_{ij}}{L_i} \right)^2 
+ \frac{\alpha_{ij}(L_i - \alpha_{ij})}{L_i^2(L_i + 1)} \right].
\]
Minimizing \(\mathcal{L}_m\) encourages the expected probabilities to approximate the ground truth while reducing uncertainty.

However, \(\mathcal{L}_m\) alone does not ensure that evidence for negative matches diminishes. To address this, we introduce a Kullback-Leibler (KL) divergence term that penalizes excessive evidence for incorrect targets. We define:
\[
\tilde{\boldsymbol{\alpha}}_i = \mathbf{y}_i + (1 - \mathbf{y}_i) \odot \boldsymbol{\alpha}_i,
\]
and measure the divergence from a uniform Dirichlet distribution \(D(\mathbf{p}_i|\mathbf{1})\):
\[
\mathcal{L}_{\text{KL}}(\boldsymbol{\alpha}_i, \mathbf{y}_i) 
= \text{KL}\big(D(\mathbf{p}_i | \tilde{\boldsymbol{\alpha}}_i) \parallel D(\mathbf{p}_i | \mathbf{1})\big).
\]
This term encourages the model to reduce unwarranted evidence for mismatched pairs, enhancing robustness against noisy matches.

For an image query \(I_i\), the evidential loss is:
\[
\mathcal{L}_e^{i2t}(I_i, l_i) = \mathcal{L}_m(\boldsymbol{\alpha}_i^{i2t}, l_i) + \lambda_2 \mathcal{L}_{\text{KL}}(\boldsymbol{\alpha}_i^{i2t}, l_i),
\]
where \(\lambda_2\) is a balance factor. Similarly, we define \(\mathcal{L}_e^{t2i}(T_i, l_i)\) for text-to-image retrieval. Combining them yields a bidirectional evidential loss:
\[
\mathcal{L}_{e}(I_i, T_i, l_i) = \mathcal{L}_e^{i2t}(I_i, l_i) + \mathcal{L}_e^{t2i}(T_i, l_i).
\]

In essence, this evidential learning framework provides a principled way to handle uncertain and noisy matches in cross-modal retrieval. By treating retrieval as a classification problem under a Dirichlet prior, we can jointly optimize for accuracy and robustness, ensuring that the model not only fits the correct matches but also avoids overconfidence in incorrect ones.

\subsection{Dynamic Softmax Hinge Loss}

Conventional hinge-based losses \cite{faghri2017vse} often consider all negative samples in a mini-batch, which can amplify errors when noisy correspondences are present. This overemphasis on all negatives may reduce model robustness and lead to unstable training. To mitigate this issue, Qin et al.~\cite{qin2022deep} proposed the Robust Dynamic Hinge (RDH) loss, focusing only on the single hardest negative. While this approach improves stability under noise, it overlooks other potentially informative negatives, limiting the model’s ability to learn from the broader negative distribution.

To achieve a more balanced and stable training process, we propose the Dynamic Softmax Hinge (DSH) loss, which adaptively utilizes a controlled subset of hard negatives rather than focusing solely on the single hardest one. By doing so, DSH maintains robustness against noise while leveraging a richer set of negative examples. This encourages a more comprehensive and effective representation learning process.

\begin{align}
\text{\tiny$\displaystyle
\mathcal{L}_h(I, T) = \frac{1}{n} \Bigg[  \sum_{j=1}^n \left[\gamma - S(I, T) + \tau\cdot\log\left(\sum_{j=1}^N\exp\left(\frac{\mathbf{S(I, \hat{T}_j)}}{\tau}\right)\right)\right]_+ \nonumber
$} \\
\text{\tiny$\displaystyle
+ \sum_{j=1}^n \left[\gamma - S(I, T) + \tau\cdot\log\left(\sum_{j=1}^N\exp\left(\frac{\mathbf{S(\hat{I}_j, T)}}{\tau}\right)\right)\right]_+ \Bigg].
$}
\end{align}

where \( [x]_+ = \max(x, 0) \), \( n \) is the dynamically adjusted number of hardest negatives, and \( \gamma \) is the margin. The value of \( n \) decreases dynamically during training:

\begin{equation}
    n = \max(\lceil K - \eta \cdot \text{Step} \rceil, \mu),
\end{equation}

where \( K \) is the mini-batch size, \( \eta \) is the annealing coefficient, \( \text{Step} \) is the current training step, and \( \mu \) is the lower bound of \( n \).

\subsection{Cross-Modal Alignment}

Triplet Ranking Loss (TRL), introduced by Schroff et al. in 2015, has been widely applied in cross-modal alignment tasks, leveraging the hardest negative samples to distinguish positive and negative pairs effectively. However, TRL suffers from performance limitations due to its insufficient focus on the broader distribution of negative samples, which can lead to suboptimal convergence and instability. To address this, Qin et al.\cite{qin2024noisy} proposed the Triplet Alignment Loss (TAL) in 2024, which relaxes the optimization from focusing solely on the hardest negative to considering all negative samples with an upper bound constraint. This relaxation reduces the risk of optimization being dominated by the hardest negatives, enhancing the stability of the training process while maintaining a comprehensive utilization of negative samples. By balancing attention to both hard and overall negative samples, TAL achieves more stable and robust training, ensuring better performance in cross-modal alignment tasks. Therefore, this paper adopts TAL as the loss function for cross-modal alignment.

For an input pair \((I_i, T_i)\) in a mini-batch \(x\), TAL is defined as:

\begin{align}
\text{\tiny$\displaystyle \mathcal{L}_{\text{TAL}}(I_i, T_i) =  \left[m - S^+_{i \to t}(I_i) + \tau \log\left( \sum_{j=1}^K \exp\left(\frac{S(I_i, T_j)}{\tau}\right)\right)\right]^+ $} \nonumber \\
\text{\tiny$\displaystyle + \left[m - S^+_{t \to i}(T_i) + \tau \log\left( \sum_{j=1}^K  \exp\left(\frac{S(I_j, T_i)}{\tau}\right)\right)\right]^+$}
\end{align}

where \(m\) is a positive margin coefficient, \(\tau\) is a temperature coefficient to control hardness, \(S(I_i, T_j) \in \{S^b_{ij}, S^t_{ij}\}\), \([x]^+ \equiv \max(x, 0)\), \(\exp(x) \equiv e^x\), and \(K\) is the size of \(x\). From Lemma 1, as \(\tau \to 0\), TAL approaches TRL and focuses more on hard negatives. Since multiple positive pairs from the same identity may appear in the mini-batch, 

\[
S^+_{i \to t}(I_i) = \sum_{j=1}^K \alpha_{ij} S(I_i, T_j)
\]

is the weighted average similarity of positive pairs for image \(I_i\).

\subsection{Overall Loss Function}
DURA aims to enhance global image-text representations in a joint embedding space by integrating multiple objectives. Specifically, it utilizes an evidential loss \((\mathcal{L}_{e})\) for handling noisy correspondences, a Dynamic Softmax Hinge loss \((\mathcal{L}_{h})\) for controlling the difficulty of negative samples, and a Triplet Alignment Loss \((\mathcal{L}_{\mathrm{TAL}})\) for stable and comprehensive cross-modal alignment. Together, these losses foster fine-grained feature interaction, uncertainty modeling, and robust image-text matching.

We train DURA in an end-to-end fashion, combining all components into a single optimization objective:
\begin{equation}
\mathcal{L}_{\text{total}} =  \mathcal{L}_{e}(I, T, l) + \mathcal{L}_{h}(I, T) +\mathcal{L}_{\text{TAL}}(I, T)
\end{equation}

\section{Experiments}

In this section, we evaluate our DURA framework on three widely used text-to-image person search datasets—CUHK-PEDES, ICFG-PEDES, and RSTPReid—under various noise conditions to demonstrate its robustness.

\subsection{Datasets and Performance Measurements}

\textbf{CUHK-PEDES}\cite{12} is the first and most widely used dataset for text-to-image person retrieval, containing 40,206 images and 80,412 textual descriptions for 13,003 unique identities. Following the official data split, the dataset is divided into a training set with 11,003 identities comprising 34,054 images and 68,108 textual descriptions; a validation set containing 1,000 identities with 3,078 images and 6,158 descriptions; and a test set also featuring 1,000 identities with 3,074 images and 6,156 descriptions. The average length of each textual description is 23 words, providing detailed visual cues for the retrieval task.

\textbf{ICFG-PEDES}\cite{23} comprises 54,522 images corresponding to 4,102 identities, with each image paired with a single textual description averaging 37 words. The training set includes 34,674 image-text pairs for 3,102 identities, while the test set consists of 19,848 image-text pairs representing the remaining 1,000 identities. This dataset is particularly notable for its one-to-one pairing of images and descriptions, emphasizing concise textual representations for each identity.

\textbf{RSTPReid}\cite{32} contains 20,505 images from 4,101 identities captured by 15 different cameras. Each identity is represented by 5 images taken from various viewpoints, and each image is annotated with 2 textual descriptions, each containing at least 23 words. Following the standard data split, the training set consists of 3,701 identities, while the validation and test sets each contain 200 identities. The diverse camera angles and specific textual annotations make RSTPReid a valuable resource for evaluating robust retrieval methods.

\textbf{Evaluation Metrics.} To assess performance, we use the Rank-k metrics (k=1,5,10), which measure the probability of retrieving a correct match within the top-k results when queried with a textual description. In addition, we employ mean Average Precision (mAP) and mean Inverse Negative Penalty (mINP)\cite{33}, providing a more comprehensive evaluation. Higher values for Rank-k, mAP, and mINP indicate superior retrieval performance.els.

\subsection{Implementation Details}

DURA uses CLIP-ViT-B/16 as the image encoder and the CLIP text Transformer as the text encoder. The multimodal interaction encoder employs a hidden size of 512 and 8 attention heads per layer. We apply standard data augmentations (random horizontal flip, random crop, random erasing) and a Key Feature Selector with a sampling ratio of 0.5. Images are resized to 384$\times$128 and text sequences are truncated to 77 tokens.

Training runs for 60 epochs with the Adam optimizer, starting at a learning rate of $8 \times 10^{-6}$, decayed using a cosine schedule and a 2-epoch warm-up. Following RDE, the margin $m$ in TAL is 0.1, and the temperature $\tau$ is 0.015. All experiments are conducted on a single RTX 2080 Ti GPU.

This setup ensures that our evaluations thoroughly measure DURA’s noise-resistant capabilities and effectiveness in extracting robust cross-modal features.

\begin{figure}[h]  
    \raggedright 
    \includegraphics[width=1\textwidth]{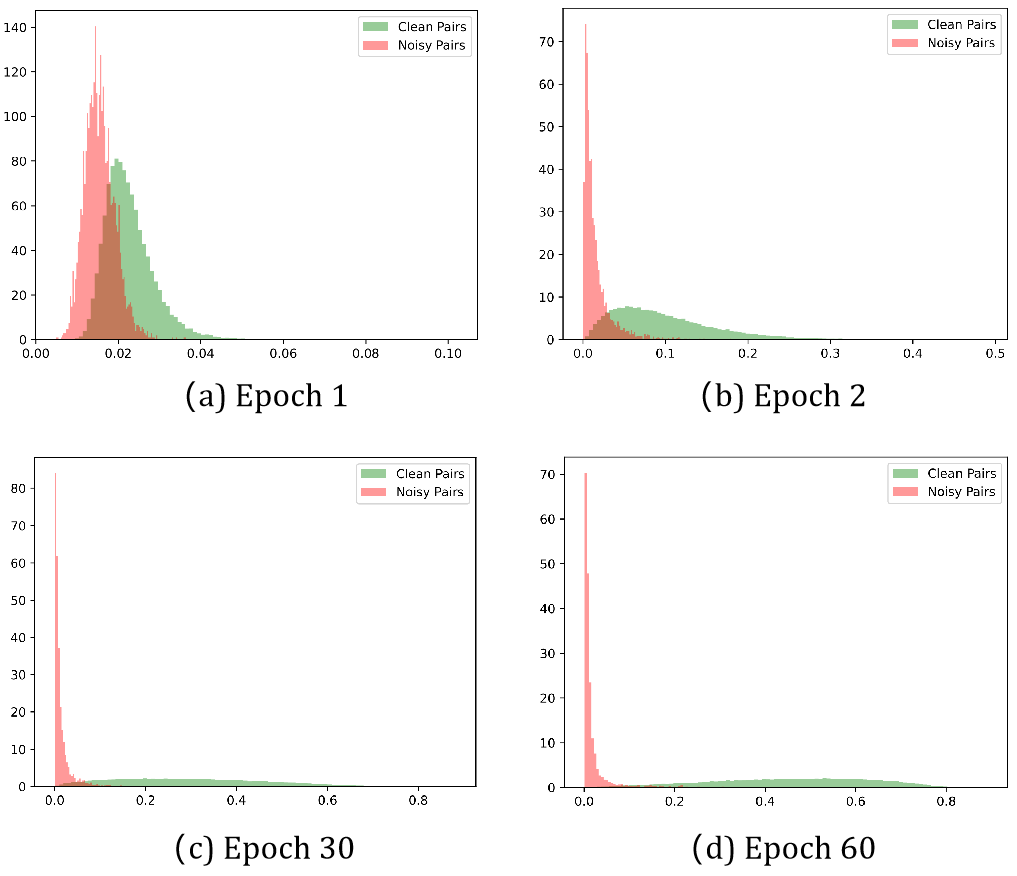} 
    \captionsetup{font=small, justification=justified, singlelinecheck=false}  
    \caption{We visualize the  evidence distribution of clean and noisy pairs at different training stages of our DURA, which is conducted on CHUKPEDES under 20\% noise. }  
    \label{fig2}  
\end{figure} 
\vspace{-10pt} 
\subsection{Comparison with Noise Correspondence}

We compare our DURA with six state-of-the-art baselines: SSAN \cite{23}, IVT \cite{18}, IRRA \cite{IRRA}, DECL \cite{qin2022deep} , RDE\cite{qin2024noisy}, and CLIP-C.

In this section, we conduct comparison experiments to evaluate the performance of our DURA model in text-to-image person search across three datasets. Our baseline is CLIP-ViT-B/16 , which achieves state-of-the-art results on all three public datasets without noisy correspondences. To verify the robustness of our framework, we inject noisy correspondences into the three datasets by randomly shuffling images for a certain percentage(i.e., 0\%, 20\%, and 50\%), and then conduct experiments on SSAN, IVT, CLIP-C, IRRA, DECL, RDE and DURA respectively. It is worth mentioning that for the sake of experimental rigour, we reran the RDE algorithm on the RTX 2080 Ti 12GB GPU, denoted by RDE*.

\begin{table*}[t]
\centering
\setlength{\tabcolsep}{2.6pt} 
\renewcommand{\arraystretch}{0.92} 

\resizebox{\textwidth}{!}{
\begin{tabular}{c|c|c|ccc|cc|ccc|cc|ccc|cc}
\hline
\multirow{2}{*}{\textbf{Noise}} &
\multirow{2}{*}{\textbf{Methods}} &
\multirow{2}{*}{\textbf{Type}} &
\multicolumn{5}{c|}{\textbf{CUHK-PEDES}} &
\multicolumn{5}{c|}{\textbf{ICFG-PEDES}} &
\multicolumn{5}{c}{\textbf{RSTPReid}} \\
\cline{4-18}
& & &
\textbf{R-1} & \textbf{R-5} & \textbf{R-10} & \textbf{mAP} & \textbf{mINP} &
\textbf{R-1} & \textbf{R-5} & \textbf{R-10} & \textbf{mAP} & \textbf{mINP} &
\textbf{R-1} & \textbf{R-5} & \textbf{R-10} & \textbf{mAP} & \textbf{mINP} \\
\hline

\multirow{7}{*}{0\%}
& SSAN & Best
& 61.37 & 80.15 & 86.73 & - & -
& 54.23 & 72.63 & 79.53 & - & -
& 43.50 & 67.80 & 77.15 & - & - \\
& IVT & Best
& 65.59 & 83.11 & 89.21 & - & -
& 56.04 & 73.60 & 80.22 & - & -
& 46.70 & 70.00 & 78.80 & - & - \\
& CFine & Best
& 69.57 & 85.93 & 91.15 & - & -
& 60.83 & 76.55 & 82.42 & - & -
& 50.55 & 72.50 & 81.60 & - & - \\
& IRRA & Best
& 73.38 & 89.93 & 93.71 & 66.13 & 50.24
& 63.46 & 80.25 & 85.82 & 38.06 & \underline{7.93}
& 60.20 & 81.30 & 88.20 & 47.17 & 25.28 \\
& DECL & Best
& 71.36 & 88.11 & 92.48 & 64.25 & 48.26
& 63.42 & 79.29 & 84.89 & 37.02 & 6.57
& 60.95 & 81.80 & 88.55 & 47.69 & 25.75 \\
& RDE$^{*}$ & Best
& \underline{75.76} & \underline{90.14} & \underline{94.18} & \underline{67.56} & \underline{51.43}
& \underline{67.68} & \underline{82.47} & \underline{87.36} & \underline{40.06} & 7.87
& \underline{65.35} & \underline{83.95} & \textbf{89.90} & \textbf{50.88} & \textbf{28.08} \\
& \textbf{DURA} & Best
& \textbf{76.14} & \textbf{90.42} & \textbf{94.25} & \textbf{67.68} & \textbf{51.57}
& \textbf{67.88} & \textbf{82.67} & \textbf{87.66} & \textbf{40.37} & \textbf{8.21}
& \textbf{66.15} & \textbf{84.55} & \underline{89.25} & \underline{50.61} & \underline{27.74} \\
\hline

\multirow{13}{*}{20\%}
& \multirow{2}{*}{SSAN} & Best
& 46.52 & 68.36 & 77.42 & 42.49 & 28.13
& 40.57 & 62.58 & 71.53 & 20.93 & 2.22
& 35.10 & 60.00 & 71.45 & 28.90 & 12.08 \\
&  & Last
& 45.76 & 67.98 & 76.28 & 40.05 & 24.12
& 40.28 & 62.68 & 71.53 & 20.98 & 2.25
& 33.45 & 58.15 & 69.60 & 26.46 & 10.08 \\
& \multirow{2}{*}{IVT} & Best
& 58.59 & 78.51 & 85.61 & 57.19 & 45.78
& 50.21 & 69.14 & 76.18 & 34.72 & \underline{8.77}
& 43.65 & 66.50 & 75.70 & 37.22 & 20.47 \\
&  & Last
& 57.67 & 78.04 & 85.02 & 56.17 & 44.42
& 48.70 & 67.42 & 75.06 & 34.44 & \textbf{9.25}
& 37.95 & 63.35 & 73.75 & 34.24 & 19.67 \\
& \multirow{2}{*}{IRRA} & Best
& 69.74 & 87.09 & 92.20 & 62.28 & 45.84
& 60.76 & 78.26 & 84.01 & 35.87 & 6.80
& 58.75 & 81.90 & 88.25 & 46.38 & 24.78 \\
&  & Last
& 69.44 & 87.09 & 92.04 & 62.16 & 45.70
& 60.58 & 78.14 & 84.20 & 35.92 & 6.91
& 54.00 & 77.15 & 85.55 & 43.20 & 22.53 \\
& \multirow{2}{*}{CLIP-C} & Best
& 66.41 & 85.15 & 90.89 & 59.36 & 43.02
& 55.25 & 74.76 & 81.32 & 31.09 & 4.94
& 54.45 & 77.80 & 86.70 & 42.58 & 21.38 \\
&  & Last
& 66.10 & 86.01 & 91.02 & 59.77 & 43.57
& 55.17 & 74.58 & 81.46 & 31.12 & 4.97
& 53.20 & 76.25 & 85.40 & 41.95 & 21.95 \\
& \multirow{2}{*}{DECL} & Best
& 70.29 & 87.04 & 91.93 & 62.84 & 46.54
& 61.95 & 78.36 & 83.88 & 36.08 & 6.25
& 61.75 & 80.70 & 86.90 & 47.70 & 26.07 \\
&  & Last
& 70.08 & 87.20 & 92.14 & 62.86 & 46.63
& 61.95 & 78.36 & 83.88 & 36.08 & 6.25
& 60.85 & 80.45 & 86.65 & 47.34 & 25.86 \\
& \multirow{2}{*}{RDE$^{*}$} & Best
& 74.46 & 89.10 & \underline{93.63} & 66.13 & 49.66
& \underline{66.54} & 81.70 & 86.70 & 39.08 & 7.55
& 64.45 & 83.50 & \textbf{90.00} & 49.78 & 27.43 \\
&  & Last
& 74.53 & 89.23 & 93.55 & 66.13 & 49.63
& 66.51 & 81.70 & 86.71 & 39.09 & 7.56
& \underline{63.85} & \textbf{83.85} & 89.45 & \underline{50.27} & \textbf{27.75} \\
& \multirow{2}{*}{\textbf{DURA}} & Best
& \textbf{75.04} & \textbf{89.74} & \textbf{93.66} & \textbf{66.81} & \textbf{50.48}
& \textbf{66.62} & \textbf{81.96} & \textbf{86.86} & \textbf{39.53} & 7.72
& \textbf{65.05} & \underline{82.75} & 89.25 & 50.11 & 27.01 \\
&  & Last
& \underline{74.60} & \underline{89.44} & 93.45 & \underline{66.43} & \underline{50.23}
& 66.53 & \underline{81.72} & \underline{86.72} & \underline{39.34} & 7.68
& \underline{64.90} & 83.20 & \underline{89.50} & \textbf{50.42} & \underline{27.50} \\
\hline

\multirow{13}{*}{50\%}
& \multirow{2}{*}{SSAN} & Best
& 13.43 & 31.74 & 41.89 & 14.12 & 6.91
& 18.83 & 37.70 & 47.43 & 9.83 & 1.01
& 19.40 & 39.25 & 50.95 & 15.95 & 6.13 \\
&  & Last
& 11.31 & 28.07 & 37.90 & 10.57 & 3.46
& 17.06 & 37.18 & 47.85 & 6.58 & 0.39
& 14.10 & 33.95 & 46.55 & 11.88 & 4.04 \\
& \multirow{2}{*}{IVT} & Best
& 50.49 & 71.82 & 79.81 & 48.85 & 36.60
& 43.03 & 61.48 & 69.56 & 28.86 & 6.11
& 39.70 & 63.80 & 73.95 & 34.35 & 18.56 \\
&  & Last
& 42.02 & 65.04 & 73.72 & 40.49 & 27.89
& 36.57 & 54.83 & 62.91 & 24.30 & 5.08
& 28.55 & 52.05 & 62.70 & 26.82 & 13.97 \\
& \multirow{2}{*}{IRRA} & Best
& 62.41 & 82.23 & 88.40 & 55.52 & 38.48
& 52.53 & 71.99 & 79.41 & 29.05 & 4.43
& 56.65 & 78.40 & 86.55 & 42.41 & 21.05 \\
&  & Last
& 42.79 & 64.31 & 72.58 & 36.76 & 21.11
& 39.22 & 60.52 & 69.26 & 19.44 & 1.98
& 31.15 & 55.40 & 65.45 & 23.96 & 9.67 \\
& \multirow{2}{*}{CLIP-C} & Best
& 64.02 & 83.66 & 89.38 & 57.33 & 40.90
& 51.60 & 71.89 & 79.31 & 28.76 & 4.33
& 53.45 & 76.80 & 85.50 & 41.43 & 21.17 \\
&  & Last
& 63.97 & 83.74 & 89.54 & 57.35 & 40.88
& 51.49 & 71.99 & 79.32 & 28.77 & 4.37
& 52.35 & 76.35 & 85.25 & 40.64 & 20.45 \\
& \multirow{2}{*}{DECL} & Best
& 65.22 & 83.72 & 89.28 & 57.94 & 41.39
& 57.50 & 75.09 & 81.24 & 32.64 & 5.27
& 56.75 & 80.55 & 87.65 & 44.53 & 23.61 \\
&  & Last
& 65.09 & 83.58 & 89.26 & 57.89 & 41.35
& 57.49 & 75.10 & 81.23 & 32.63 & 5.26
& 55.00 & 80.50 & 86.50 & 43.81 & 23.31 \\
& \multirow{2}{*}{RDE$^{*}$} & Best
& \textbf{71.33} & \textbf{87.41} & \textbf{91.81} & \underline{63.50} & \underline{47.36}
& 63.76 & 79.53 & \textbf{84.91} & 37.38 & 6.80
& \underline{62.85} & \underline{83.20} & \underline{89.15} & \underline{47.67} & 23.97 \\
&  & Last
& \underline{71.25} & \underline{87.39} & 91.76 & \textbf{63.59} & \textbf{47.50}
& 63.76 & 79.53 & \textbf{84.91} & 37.38 & 6.80
& \underline{62.85} & \underline{83.20} & \underline{89.15} & \underline{47.67} & 23.96 \\
& \multirow{2}{*}{\textbf{DURA}} & Best
& 70.89 & 87.21 & \underline{91.78} & 63.13 & 46.68
& \textbf{64.08} & \textbf{79.87} & \underline{84.6} & \textbf{37.57} & \textbf{7.06}
& \textbf{62.95} & \textbf{83.55} & \textbf{89.45} & \textbf{47.92} & \textbf{25.55} \\
&  & Last
& 70.84 & 87.04 & \underline{91.78} & \textbf{63.59} & 46.85
& \underline{63.85} & \underline{79.63} & 84.58 & \underline{37.55} & \underline{7.02}
& 61.85 & 82.80 & 88.65 & 47.45 & \underline{25.04} \\
\hline
\end{tabular}
}

\caption{Horizontal comparison under different noise rates on CUHK-PEDES, ICFG-PEDES, and RSTPReid. ``Best'' and ``Last'' follow the same protocol as in the main text. R-1/R-5/R-10 are retrieval accuracies (\%). The best and second-best results are highlighted in \textbf{bold} and \underline{underlined}, respectively.}
\label{tab:noise_horizontal_compact}
\end{table*}

\subsubsection{Comparisons on CUHK-PEDES} 

We first evaluate our proposed method DURA on the most common benchmark CUHK-PEDES with noisy correspondences. As shown in Table \ref{tab:noise_horizontal_compact}, the experimental results show that our DURA could improve the robustness of baselines against noisy correspondence.  
Particularly, at a noise level of 0.2, the five scores achieved by DURA are 75.04\%, 89.74\%, 93.66\%, 66.81\% and 50.48\%, significantly outperforming other six methods. This indicates that the retrieval performance of DURA is markedly more resilient under noise conditions. Overall, the proposed DURA method outperforms the other six methods in the text-to-image person search task on the CUHK-PEDES dataset, demonstrating superior performance across both low and high noise levels. Notably, the DURA method exhibits less performance degradation under high noise conditions, which underscores its enhanced robustness when handling complex scenarios.

\subsubsection{Comparisons on ICFG-PEDES} 
Table \ref{tab:noise_horizontal_compact} presents the experimental results on the ICFG-PEDES 1K test set, demonstrating that DURA achieves superior performance under both low and high noise conditions. Compared to six baseline methods, DURA consistently outperforms across all evaluation metrics (Rank-1, Rank-5, Rank-10, mAP, and mINP), particularly at noise levels of 0.2 and 0.5. Under a noise level of 0.5, DURA shows a significant improvement in the Rank-1 metric, achieving up to a 12\% gain over the baselines. These results highlight the robustness and effectiveness of DURA in addressing noisy correspondence and maintaining high retrieval performance in challenging scenarios.

\subsubsection{Comparisons on RSTPReid}  Table \ref{tab:noise_horizontal_compact} shows retrieval results on the RSTPReid dataset. From the experimental results, we can see that our DURA performs extremely well in the high noise case, with all five metrics located at the highest, 62.94\%, 83.55\%, 89.45\%, 47.92\% and 25.55\%, respectively.This indicates that DURA adapts well to the high noise case on the RSTPReid dataset. It is worth noting that RDE also performs well in high noise adaptation on this dataset, second only to our DURA.

\subsection{Ablation Study}
In this section, we perform an ablation study on the CUHK-PEDES dataset with 20\% noise, and we adopt the CLIP-ViT-B/16 model as the baseline, aiming to analyse the effectiveness of each component in the DURA framework. After analysing the results of Table \ref{tab4}, we obtain the following conclusions: (1) By comparing No.0 and No.1, it is found that the addition of TAL alone results in an increase of +4.04\%, +1.59\%, +0.63\%, +2.97\%, and +2.75\% for Rank-1, Rank-5, Rank-10, mAP, and mINP, respectively. (2) Through the comparison of No.1 and No.4, it is found that the addition of KFS on top of TAL results in a large increase in all metrics, which is similarly demonstrated in the comparison of No.2 and No.6, and No.3 and No.5. (3) By comparing No.4 with No.5 and No.4 with No.6, it is found that the addition of $\mathcal{L}_{e}$ and $\mathcal{L}_{h}$ on top of the addition of TAL and KFS, respectively, both lead to a partial improvement in performance. (4) By comparing No.7 with other data, we find that DURA achieves the best results in all five metrics, with Rank-1, Rank-5, Rank-10, mAP and mINP reaching 75.04\%, 89.74\%, 93.66\%, 66.81\% and 50.48\%, respectively. This demonstrates that the components of our DURA is effectively stacked with superior stability and robustness.


\begin{table}[htbp]
    \centering
    \setlength{\tabcolsep}{4pt}        
    \renewcommand{\arraystretch}{1.08} 

    \small 
    \begin{tabular}{c|l|ccccc}
        \hline
        \textbf{No.} & \textbf{Methods} & \textbf{R-1} & \textbf{R-5} & \textbf{R-10} & \textbf{mAP} & \textbf{mINP} \\
        \hline
        0 & Baseline & 66.41 & 85.15 & 90.89 & 59.36 & 43.02 \\
        1 & +TAL & 70.45 & 86.74 & 91.52 & 62.33 & 45.77 \\
        2 & +TAL+$\mathcal{L}_{h}$ & 71.05 & 87.15 & 91.80 & 63.26 & 46.89 \\
        3 & +TAL+$\mathcal{L}_{e}$ & 70.74 & 87.22 & 91.85 & 63.25 & 46.85 \\
        4 & +TAL+KFS & 73.44 & 88.87 & 92.32 & 65.24 & 48.69 \\
        5 & +TAL+KFS+$\mathcal{L}_{e}$ & 74.58 & 88.97 & 92.73 & 66.09 & 49.81 \\
        6 & +TAL+KFS+$\mathcal{L}_{h}$ & 74.08 & 88.74 & 92.52 & 65.96 & 49.09 \\
        \hline
        7 & +TAL+KFS+$\mathcal{L}_{e}$+$\mathcal{L}_{h}$ & \textbf{75.04} & \textbf{89.74} & \textbf{93.66} & \textbf{66.81} & \textbf{50.48} \\
        \hline
    \end{tabular}

    \captionsetup{font=small,justification=justified,singlelinecheck=false}
    \caption{Ablation study on the CUHK-PEDES dataset.}
    \label{tab4}
\end{table}
\subsection{Robustness Study}
In this section, we present visualization results obtained during cross-modal training to demonstrate the robustness and effectiveness of our method. As illustrated in Figure 3, it is evident that our DURA not only delivers outstanding performance in the presence of noise but also effectively mitigates noise overfitting.

\begin{figure}[htbp]  
    \raggedright 
    \includegraphics[width=1
    \textwidth]{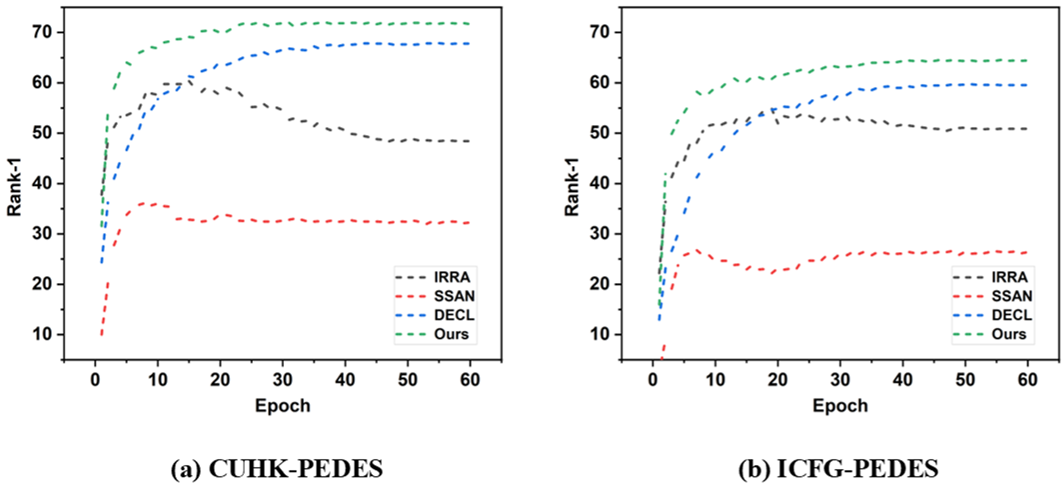} 
    \captionsetup{font=small, justification=justified, singlelinecheck=false}  
    \caption{Test performance (Rank-1) versus epochs on the CHUKPEDES and ICFG-PEDES datasets with 50\% noise.}  
    \label{fig3}  
\end{figure} 

\section{Conclusion}
In this work, we present the Dynamic Uncertainty and Relational Alignment (DURA) framework, aimed at improving  text-based person
search under noisy conditions. DURA leverages a Key Feature Selector (KFS) and a Dynamic Softmax Hinge Loss (DSH-Loss) to robustly handle noisy training pairs, while its cross-modal evidential learning distinguishes clean from noisy data. Experiments on CUHK-PEDES, ICFG-PEDES, and RSTPReid confirm that DURA consistently outperforms existing methods, especially in high-noise scenarios, demonstrating its practical effectiveness when data quality is limited.

\bibliography{main}
\bibliographystyle{plain} 
\end{document}